\documentclass[letterpaper,10pt,conference]{ieeeconf}
\IEEEoverridecommandlockouts   
\usepackage[english]{babel}
\usepackage{amsmath}
\usepackage{lipsum} 
\usepackage[T1]{fontenc}
\usepackage{newtxtext}
\usepackage[cmintegrals]{newtxmath}
\usepackage{bm} 
\usepackage[linesnumbered,ruled,vlined]{algorithm2e}
\usepackage[utf8]{inputenc}
\usepackage{graphicx}
\usepackage{graphics}
\usepackage{subcaption}
\usepackage{siunitx}
\usepackage{multirow}
\usepackage{float}
\usepackage{caption}
\usepackage{gensymb}
\usepackage{hyperref}
\usepackage{epstopdf}
\usepackage{enumerate}
\usepackage{multicol,multirow}
\usepackage{booktabs}
\usepackage{cite}
\usepackage{balance}
\usepackage{tikz,xcolor}
\usepackage{academicons}
\DeclareCaptionLabelSeparator{periodspace}{.\quad}
\usepackage{svg}
\definecolor{orcidlogocol}{HTML}{A6CE39}
\graphicspath{{figures/}} 
\definecolor{lime}{HTML}{A6CE39}
\DeclareRobustCommand{\orcidicon}{%
	\begin{tikzpicture}
	\draw[lime, fill=lime] (0,0) 
	circle [radius=0.16] 
	node[white] {{\fontfamily{qag}\selectfont \tiny ID}};
	\draw[white, fill=white] (-0.0625,0.095) 
	circle [radius=0.007];
	\end{tikzpicture}
	\hspace{-2mm}
}

\foreach \x in {A, ..., Z}{
	\expandafter\xdef\csname orcid\x\endcsname{\noexpand\href{https://orcid.org/\csname orcidauthor\x\endcsname}{\noexpand\orcidicon}}
}

\begin{document}
\bstctlcite{IEEEexample:BSTcontrol}
	\title{\LARGE\bf Robust High-Transparency Haptic Exploration for Dexterous Telemanipulation}
	\author{Keyhan~Kouhkiloui~Babarahmati, Carlo~Tiseo, Quentin~Rouxel,\\ Zhibin~Li, and~Michael~Mistry
		\thanks{All authors are with the School of Informatics, University of Edinburgh. Email: keyhan.kouhkiloui@ed.ac.uk, carlo.tiseo@ed.ac.uk.}
		\thanks{
This work has been supported by EPSRC UK RAI Hub ORCA (EP/R026173/1), the Future AI and Robotics for Space (EP/R026092/1), National Centre for Nuclear Robotics (NCNR EPR02572X/1) and THING project in the EU Horizon 2020 (ICT-2017-1).}
	}
	\thispagestyle{empty}
\fbox{
\parbox{\textwidth}{
© 2021 IEEE. Personal use of this material is permitted.  Permission from IEEE must be obtained for all other uses, in any current or future media, including reprinting/republishing this material for advertising or promotional purposes, creating new collective works, for resale or redistribution to servers or lists, or reuse of any copyrighted component of this work in other works.}}
\newpage

\maketitle
	
\begin{abstract}
Robotic teleoperation provides human-in-the-loop capabilities of complex manipulation tasks in dangerous or remote environments, such as for planetary exploration or nuclear decommissioning.
This work proposes a novel telemanipulation architecture using a passive Fractal Impedance Controller (FIC), which does not depend upon an active viscous component for guaranteeing stability. Compared to a traditional impedance controller in ideal conditions (no delays and maximum communication bandwidth), our proposed method yields higher transparency in  interaction and demonstrates superior dexterity and capability in our telemanipulation test scenarios. We also validate its performance with extreme delays up to \SI{1}{\second} and communication bandwidths as low as \SI{10}{\hertz}. All results validate a consistent stability when using the proposed controller in challenging conditions, regardless of operator expertise.
\end{abstract}
	
	
\IEEEpeerreviewmaketitle
	
\section{INTRODUCTION} \label{sec:Introduction}
Exploration robots have been deployed in space, such as to the Moon and Mars. Recent development in space robotics has focused on more dexterous telemanipulation \cite{wen2020force} that can enhance robot-driven exploratory missions and relieve astronauts from dangerous tasks. For example, the European Space Agency's METERON (Multi-Purpose End-To-End Robotic Operation Network) project aims to demonstrate the operational ability to teleoperate planetary robots from in-orbit crewed stations \cite{schmaus2018preliminary, schmaus2019continued}. Tasks such as collection and on-site analysis of geological samples, as well as assembly and construction of planetary structures are strongly demanded. Similar capabilities are also needed on earth where hazardous environments (e.g.\ nuclear, offshore, underwater) prevent direct human intervention. These tasks require dexterous and delicate manipulation involving changes of physical contact and coordination of two arms \cite{sun2020learning}, which are challenging to model or fully automate in unknown environments. Human-in-the-loop guidance based on multimodal sensory interfaces, especially haptics, can enhance the capability but will be complicated by signal latency and low-bandwidth communication inherent within these extreme environments.

\begin{figure}[!tb]
\centering
\includegraphics[width = \columnwidth]{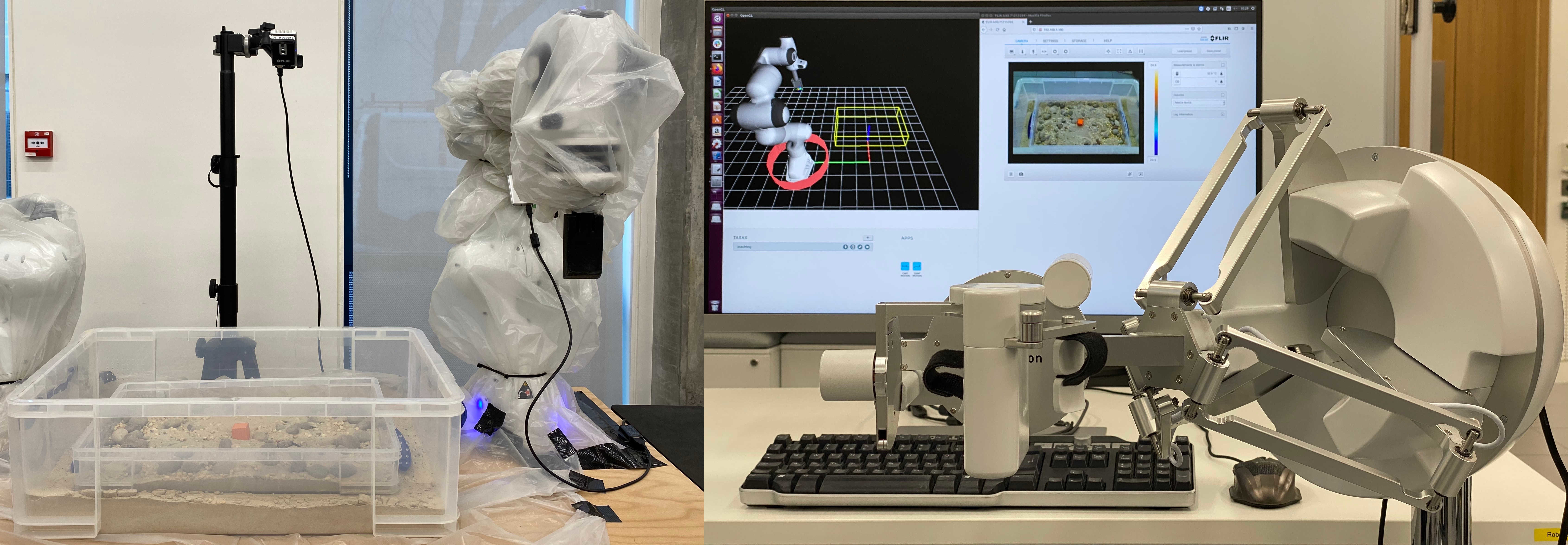}
\caption{Experimental setup: The replica robot is a Panda 7-DoF Manipulator Arm with mock terrestrial planet environment, and the master setup is a Sigma.7 with a GUI.}
\label{fig:bilateralTelemanipulationSetup}
\end{figure}

\begin{figure*}[!htb]
\vspace{1mm}
\centering
\includegraphics[width=0.9\linewidth]{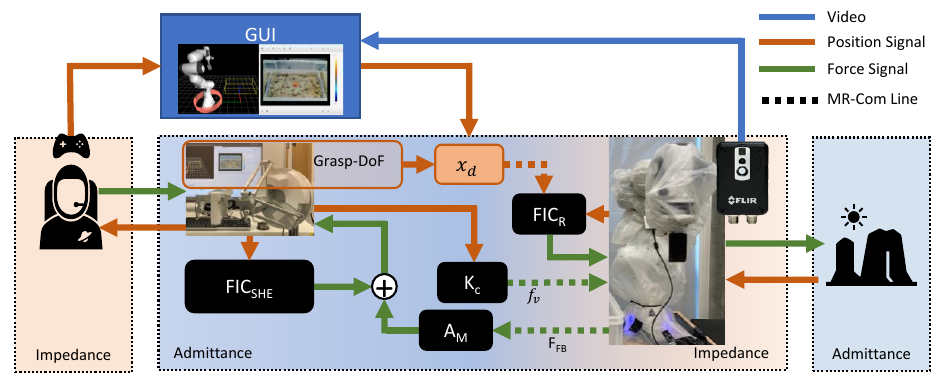}
\caption{{Multimodal Teleoperation System Overview}: The operator of the master device (Sigma.7, Force Dimension) drives the replica (Panda, Franka Emika AG) by applying a virtual force at its end-effector ($f_\text{v}$), which is equal to the product between master pose ($x_\text{M}$) and a constant stiffness ($K_\text{c}$). Interaction force/torque feedback at the end-effector ($F_{\text{FB}}$) measured by an Optoforce HEX-70-CH-4000N is relayed back to the operator via an admittance gain module (A$_\text{M}$). The operator can adjust the replica end-effector's reference pose ($x_\text{d}$) by activating the master device using the grasp joint or GUI. The GUI also provides visual feedback from an RGB camera (FLIR-AX8) and a digital environment model. Dashed lines (MR-Com) indicate signals affected by the delay and low-bandwidth communication in our experiments.}
\label{fig:proposedCtrlBlocks}
\end{figure*}

Port-Hamiltonian controllers, such as impedance \cite{impCtrl,Hogan2014} and admittance control \cite{li2016compliance}, have been identified as good candidates to enhance the interaction capabilities of robots deployed in such environments. 
These controllers are designed to drive the robot as an equivalent mass-spring-damper system.
However, the coupling of the robot dynamics with the environment poses a threat to both their stability and robustness, if the robot is not able to track the energy exchanged during interaction \cite{forceImpTrajLearning,varImpCtrlReinforcementLearApproach,varImpCtrlBasedOnHumanStiffEstimation}.
Furthermore during bilateral telemanipulation, information delay may be highly detrimental to the interaction, increasing the risk of unstable behaviour and reducing system robustness.

A known solution for maintaining stability is guaranteeing the passivity of the system. Based on this theory, any passive interaction with a passive system is guaranteed to be stable, and moreover any combination of passive systems is also guaranteed to be passive \cite{van2000l2}. Furthermore, passive systems have been shown to be robust to information delays and reduced communication bandwidth \cite{hulin2013passivity,babarahmati2019fractal,stramigioli2005}, which is essential for teleoperation.
Passivity-based approaches have been implemented to derive necessary conditions for stability with respect to the energy flow of the system \cite{ryu2004stable}.
However, the traditional passive impedance control has been found only feasible under constrained conditions, which are determined by the aforementioned coupling of the robot and environmental dynamics \cite{hogan2018impedance}. 
Therefore, \textit{Virtual energy tanks} have been proposed to achieve dexterous manipulation via the passivity framework \cite{ferraguti2015energy, positionDriftCompensation,bilateralTeleManWithTimeDelays, portBasedAsymptoticCurveTracking, combiningEnergyAndPowerBasedSafety, safetyAndEnergyAwareImpCtrl}.
Energy tank controllers enforce passivity of an impedance control by tracking the energy exchange between the system and the environment. 
The non-conservative energy exchanges are tracked using virtual springs (i.e., energy tank).
This allows the system to use stored energy to guarantee that the extracted energy is less than or equal to injected energy \cite{stabilityConsiderationsVariImpCtrl,unifiedPassivityBased,passivationOfProjectionBased}.
Thus, they retain a ``memory'' of the interaction and return to the environment an energy equal or lower to the received amount. They can be used to implement passive variable impedance controllers \cite{ferraguti2013tank, ferraguti2015energy}. 
However, the tank-based controllers are sensitive to the degradation of energy tracking \cite{passivationOfProjectionBased}, which may occur in dexterous teleoperation with feedback delay and uncertainty in environmental interaction. 

The Fractal Impedance Controller (FIC) is a novel passive framework recently proposed in \cite{babarahmati2019fractal}. Differently from other variable impedance controllers, it does not rely upon an active viscous component (i.e.\ damping) for velocity tracking and stability. 
Instead, it achieves asymptotic stability via an anisotropic behaviour between divergence and convergence (a hysteresis cycle).
The anisotropy is controlled by a switching controller that introduces an additional impedance when a velocity zero-crossing is detected. 
Furthermore, the controller has a maximum exertable force that can be calibrated based on system characteristics once the desired stiffness profile during divergence has been chosen. All other control parameters are automatically derived.
Differently from other controllers, it does not rely on the tracking of non-conservative forces for passivity, making it extremely suitable for applications with interaction uncertainties and information delays.

This research uses the setup shown in \autoref{fig:bilateralTelemanipulationSetup} to validate the control architecture in \autoref{fig:proposedCtrlBlocks} and compares the interaction behaviour of the fractal impedance controller with a traditional impedance controller in teleoperation tasks with low-bandwidth communication and signal delay. We also consider other passive controllers for this application (i.e., energy tank). However, their stability conditions are based on the trading-off of tracking accuracy in order to guarantee stability which is not ideal for dexterity and transparency \cite{passivationOfProjectionBased,ferraguti2015energy,bilateralTeleManWithTimeDelays,minelli2019energy}. The fractal impedance controller does not require such a trade-off. 

The contribution of this paper is the development of a passive teleoperation framework based on the Fractal Impedance Controller (FIC) proposed in \cite{babarahmati2019fractal}, which being passive guarantees the stability of the system even in the presence of delays or reduced bandwidth. We validate the proposed architecture both with and without significant time delay and reduced bandwidth in the communication channel between the master and the replica, and quantify the performance degradation related to long distance telecommunication. Lastly, we also compare the dexterity of an expert user in 5 tasks in local operations, using both the FIC and a traditional impedance controller on the replica. 

The rest of the paper is organised as follows: In section \ref{sec:Methodology}, the design of the proposed method for passive bilateral telemanipulation is presented. Then, section \ref{Experimental_Validation} discusses the experimental validation followed by section \ref{sec:Results}, which consists of the experimental results of the carried out experiments. In the end, discussion, the conclusion and future works are provided in section \ref{sec:Discussion}.

\section{CONTROL ARCHITECTURE} \label{sec:Methodology}
High levels of transparency between master and replica devices are essential for  dexterious interaction in telemanipulation \cite{zhou2020relationship}. Ideal transparency implies that the system acts as an identity transformation with infinite bandwidth (i.e, there is effectively no intermediary between the operator and the environment). Our proposed architecture aims at improving system transparency via FIC, which we argue is more robust to discretisation, model-errors, noise and time-delay than a traditional impedance controller.

The proposed method for bilateral haptic manipulation (\autoref{fig:proposedCtrlBlocks}) has two physical interfaces. The first interface, between the operator and the master device (Sigma.7, Force Dimension), has an admittance controller (i.e., $\dot{x}=Z^{-1}f_\text{d}$,\cite{impCtrl}) that reproduces interaction dynamics at the replica robot's end-effector. The second interface, between the replica (Panda, Franka Emika) and the environment, behaves as an impedance mimicking the operator input acquired by the master device. We also propose the use of an additional impedance controller -- FIC -- for Safety and Haptic Enhancement (FIC$_{\text{SHE}}$), generating a virtual force added to the feedback from the replica for providing additional safety to the operator, such as virtual work-space boundaries.
The master device is controlled using an admittance controller that includes the Sigma.7 gravity compensation.
The master position error ($\tilde{x}_\text{M}$) is the input to a proportional controller generating a virtual force ($f_\text{v}$) on the replica's end-effector. 
The master's gripper button allows the operator to switch to velocity control, whereby when the gripper is depressed the replica's desired position ($x_\text{Rd}$) is continuously updated by adding the current master position error to the initial value of $x_{\text{Rd}}$ until the gripper is released. 

Despite these control modalities being useful in dexterous fine interaction, they are not easy to use for large movements. Therefore, we have also implemented a keyboard based input through a GUI to modify the desired pose and autonomously drive the robot to the desired pose. Finally, the GUI is also used to provide visual feedback of the replica's workspace via a camera.

\subsection{Master's Controllers}
The master desired wrench command input for the Sigma-7's admittance controller (as defined in \cite{impCtrl}) is sum of the scaled end-effector force  ($A_\text{M}$) and the FIC$_{\text{SHE}}$:
\begin{equation} \label{eq:fractalImpCtrlMasterSide}
    W_{\text{M}} = {\text{FIC}}_{\text{SHE}} + {\text{A}}_{\text{M}}=K_{\text{M}}(\tilde{x}_{\text{M}}){\tilde{x}{_{\text{M}}}} - D_{\text{M}}{\dot{{x}}}_{\text{M}} + K_\text{A}F_{\text{FB}}
\end{equation}
where $K_{\text{M}}(\tilde{x})$ is a state dependant diagonal stiffness matrix, $D_{\text{M}}$ is the (diagonal) damping gain, $F_\text{FB}$ is the wrench measured at the replica's end-effector and $K_A$ is the scaling factor set to $1$ in this work.
\subsection{Replica's Controllers}
The replica's controller is the combination of the virtual force from the master ($J_\text{R}^{T}K_\text{c}\tilde{x}_\text{M}$), the FIC$_\text{R}$, the replica's inverse dynamics compensation and a null-space controller (in the case of a redundant manipulator):
\begin{equation} \label{eq:fractalImpCtrlSlaveSide}
      \begin{array}{rl}
          \tau_{\text{FICr}}=& J_\text{R}^{T}(K_\text{c}x_\text{M} + K_{\text{R}}(\tilde{x}_{\text{R}}){\tilde{x}}_{\text{R}} - D_{\text{R}}{\dot{{x}}}_{\text{R}}\\ 
          &+ h_\text{R}(q_\text{R},\dot{q}_\text{R})) +(I - J_\text{R}^{T}(J_\text{R}J_\text{R}^T)^{-1} J_\text{R}) {\tau}_{\text{Rnull}} \\
          &+g_\text{R}(q_\text{R}),
      \end{array}
\end{equation}
where $J_{\text{R}}$ is the end-effector Jacobian and $\tau_{\text{Rnull}} = K_{\text{Rnull}} {\tilde{q}}_\text{R} - D_{\text{R}} \dot{q}_\text{R}$ is the null space torque ($\tilde{q}_\text{R}$ is the error from the reference pose and $\dot{q}_\text{R}$ are joint velocities). $h_\text{R}(q_\text{R}) = \Lambda_\text{R} (q_\text{R})(J_\text{R} M_\text{R}^{-1}C_\text{R}(q_\text{R},\dot{q}_\text{R})\dot{q}_\text{R} - \dot{J_\text{R}}\dot{q}_\text{R})$ is the inverse dynamics compensation, where $M_\text{R}$ is the inertia matrix. 
$\Lambda_\text{R} (q) = {(J_\text{R}M_\text{R}^{-1}J_\text{R}^{T})}^{-1}$ is the task-space inertia matrix, $ C_\text{R}(q,\dot{q})$ is Coriolis's matrix, and  $g_\text{R}(q_\text{R})$ is the gravity compensation vector in joint-space. $K_c$ is the scaling coefficient for the virtual force generated by the operator, $K_{\text{R}}$ and $D_{\text{R}}$ are the diagonal matrices for stiffness and damping gains, respectively.

\subsection{Fractal Impedance Controller}
The FIC is a anisotropic passive controller described in details in \cite{babarahmati2019fractal}. It is used in both FIC$_{\text{SHE}}=K_{\text{M}}(\tilde{x}_{\text{M}}){\tilde{x}{_{\text{M}}}} - D_{\text{M}}{\dot{{x}}}_{\text{M}}$ and FIC$_{\text{R}}=K_{\text{R}}(\tilde{x}_{\text{R}}){\tilde{x}{_{\text{R}}}} - D_{\text{R}}{\dot{{x}}}_{\text{R}}$. Thus, generating the following end-effector dynamics:
\begin{equation}
    \label{FICDyn}
    \Lambda_i\ddot{x}_i-D_i\dot{x}+K_i(\tilde{x}_i)\tilde{x}_i=0, ~ i=\text{R},~\text{M}
\end{equation}
where $\Lambda_i$ is the task-space inertia, $D_i$ is a constant damping, and $K_i$ is the anisotropic state-dependant impedance profile, determined as follows:

\textbf{Divergence Phase ($sgn(\dot{\tilde{x}}_i)=sgn(\tilde{x}_i)$ or $\dot{\tilde{x}}_i=0$):}
        \begin{equation} \label{eq:KdSaturation}
            \begin{array}{l}
                K_i (\tilde{x}_i) = K_{0i} + K_{\text{v}i}(\tilde{x}_i) \ \text{where}\\
                {K_{\text{v}i}}(\tilde{x}_i) = \begin{cases}
                \displaystyle{\frac{W_{\text{max}i}}{|{\tilde{x}_i}|}} - K_{0i}, \  	& \ \text{if} \ |\tilde{x}_i| > x_{\text{B}i} \\
                \displaystyle{{e}^{({\beta_i} {\tilde{x}_i})^2}}, & \  \text{otherwise}
                \end{cases}
            \end{array}
        \end{equation}  

\textbf{Convergence Phase ($sgn(\dot{\tilde{x}})\ne sgn(\tilde{x})$):}
        \begin{equation} \label{eq:stiffnessProfileInConv}
            \left.
                \begin{array}{l}
                K_i (\tilde{x}_i) = \left( \frac{4}{\tilde{x}^2_{\text{max}i}}\right) \mathlarger{\int_0 ^{\tilde{x}_{\text{max}i}}{  K_{\text{D}i}(\tilde{x}_{i}) \tilde{x}_i~d\tilde{x}_i}} \left(\frac{0.5\tilde{x}_{\text{max}i} -\tilde{x}_i}{\tilde{x}_i}\right) 
                \end{array}
            \right.
        \end{equation}

Where $K_{0i}$ and $K_{\text{v}i}$ are the constant and the variable stiffness respectively, $W_{\text{max}i}$ is the maximum exertable wrench, $K_{\text{D}i}$ is the stiffness during divergence (\autoref{eq:KdSaturation}). $\tilde{x}_{\text{max}i}$ is the maximum position error reached during divergence, and ${\beta_i}$ controls the stiffness saturation-boundaries, which is calculated as follows: 
\begin{equation} \label{eq:KmaxCalculation}
\begin{array}{c}
K_{\text{max}i} = \displaystyle{^{W_{\text{max}i}}/_{x_{\text{B}i}}}\\
\beta_i = \displaystyle{\sqrt{^{\ln({K_{\text{max}i}})}/_{{x_{{B}i}}^2}}} 
\end{array}
\end{equation}
where $x_{\text{B}i}$ is the pose error where  $W_{\text{max}i}$ is reached. For full explanation of these equations we refer the reader to \cite{babarahmati2019fractal}.

\section{EXPERIMENTAL VALIDATION} \label{Experimental_Validation}
We compare our fractal impedance controller to a traditional impedance controller (IC) from the perspective of the operator, as shown in \autoref{fig:proposedCtrlBlocks}. The traditional impedance controller is similar to the FIC$_\text{R}$ control law (\autoref{eq:fractalImpCtrlSlaveSide}), however, it only has a constant stiffness term. For these experiments, we keep the constant stiffness terms ($K_{0i}$ for FIC$_\text{R}$) the same for both controllers. In order to keep IC stable for our experiments, we also had to set its damping term 8 times higher than FIC$_\text{R}$'s. The values for the three controllers' parameters used in this paper are reported in \autoref{tab:controller_gains}. To evaluate transparency, we conduct spectral analysis considering the force signals measured on the replica's end-effector as input and the velocity produced on the master as output. We also examine the robustness of the controller in the presence of low-bandwidth and time-delayed feedback via experiments with an expert and a naive subject to evaluate whether the stability of the proposed method relies on the operator skill. In summary, the following experiments are performed:

	\begin{table}[b]
		\caption{Controller Parameters}	
		\begin{center}
				\begin{tabular}{| c | c | c | c | c|}
					\hline 
					\multicolumn{1}{|c|}{}  & \multicolumn{2}{|c|}{Master} & \multicolumn{2}{|c|}{Replica}
					\\ \hline
					 \multicolumn{1}{|c|}{}  & \multicolumn{2}{|c|}{FIC$_\text{SHE}$}  & \multicolumn{1}{|c|}{FIC$_\text{R}$} & \multicolumn{1}{|c|}{IC}
					\\ \hline 
					$K_{\text{c}}^{\text{linear}} \SI{}{(\newton\per\meter)}$ & \multicolumn{2}{|c|}{0}  & 100 & 100\\ \hline
					$K_{\text{c}}^{\text{angular}} \SI{}{(\newton\per\radian)}$ & \multicolumn{2}{|c|}{0} & 5 & 5 \\ \hline
					$D^{\text{linear}} \SI{}{(\newton\second\per\meter)}$ & \multicolumn{2}{|c|}{2.5} & 2.5 & 20 \\ \hline
					$D^{\text{angular}} \SI{}{(\newton\second\per\radian)}$ & \multicolumn{2}{|c|}{0} & 1.25 & 1.25 \\ \hline
					$x_{B}^{\text{linear}} \SI{}{(\meter)}$ & \multicolumn{2}{|c|}{0.075} & 0.075 & NA \\ \hline
					$x_{B}^{\text{angular}} \SI{}{(\radian)}$ & \multicolumn{2}{|c|}{1.0472} & 1.0472 & NA \\ \hline
    				\end{tabular}
			\caption*{FIC: Fractal Impedance Control, IC: Impedance Control.}
		\end{center}
		\label{tab:controller_gains}
	\end{table}

\begin{figure}[h]
\centering
    \vspace{4mm}
    \begin{subfigure}[b]{0.43\columnwidth}
	    \centering
		\includegraphics[width=\textwidth]{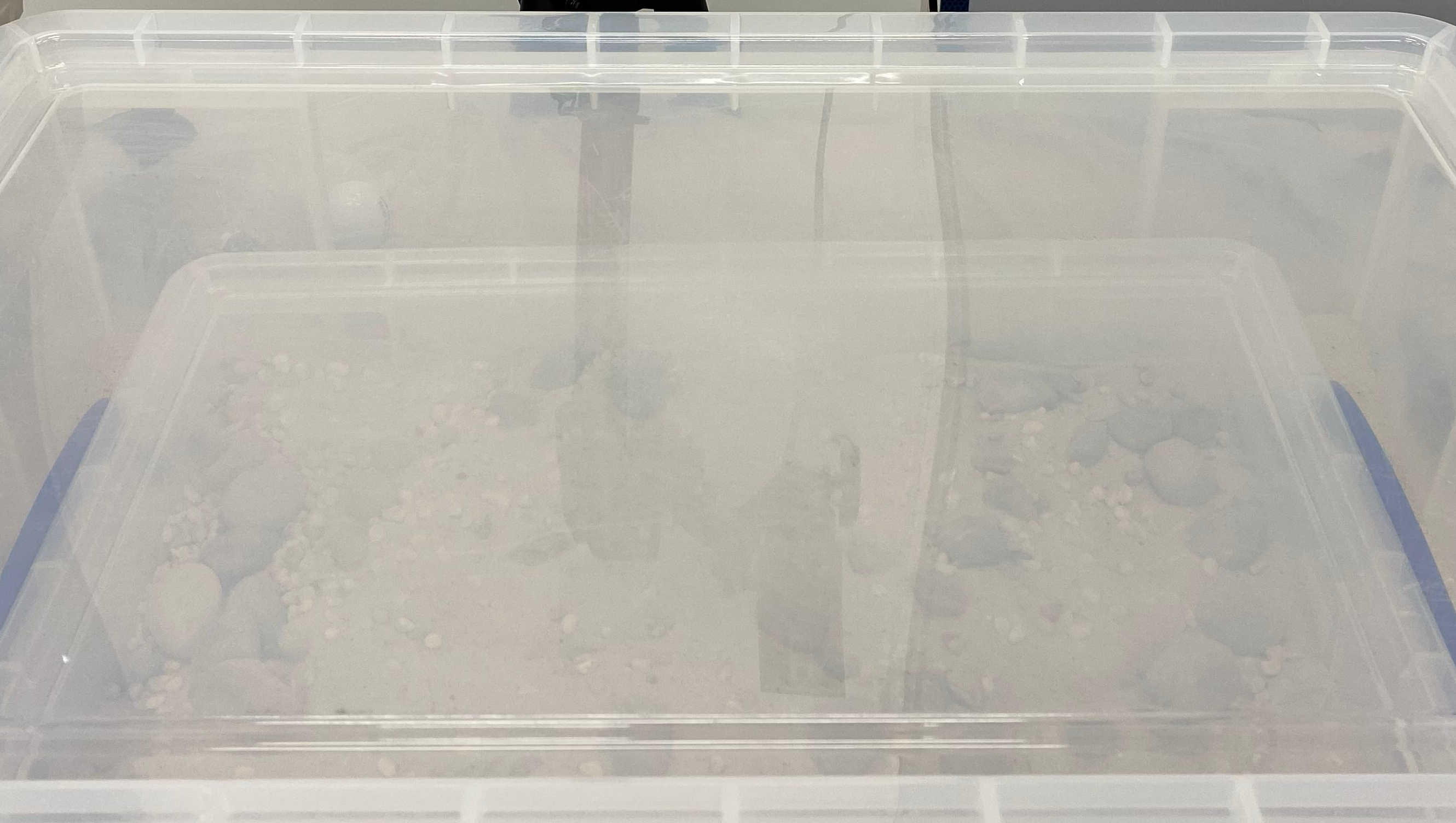}
		\caption{\label{fig:Obj_1} Object 1}
	\end{subfigure}
	~~~
	\begin{subfigure}[b]{0.185\columnwidth}
		\centering
		\includegraphics[width=\textwidth]{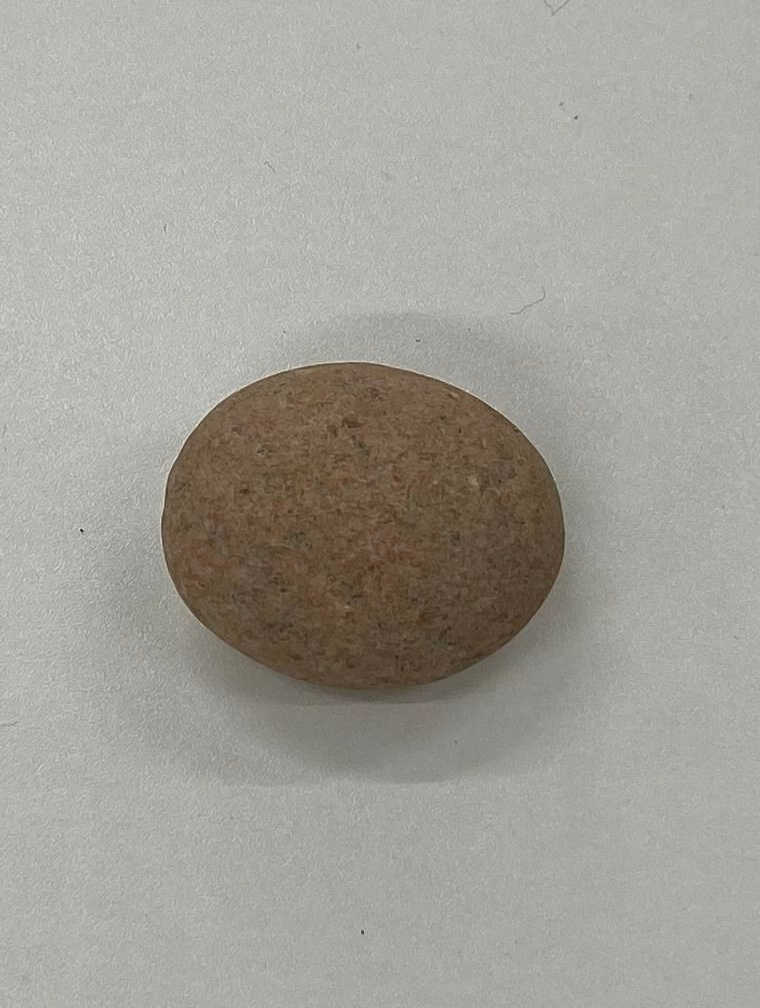}
		\caption{\label{fig:Obj_2} Object 2}
	\end{subfigure}
	~~~
	\begin{subfigure}[b]{0.28\columnwidth}
		\centering
		\includegraphics[width=\textwidth]{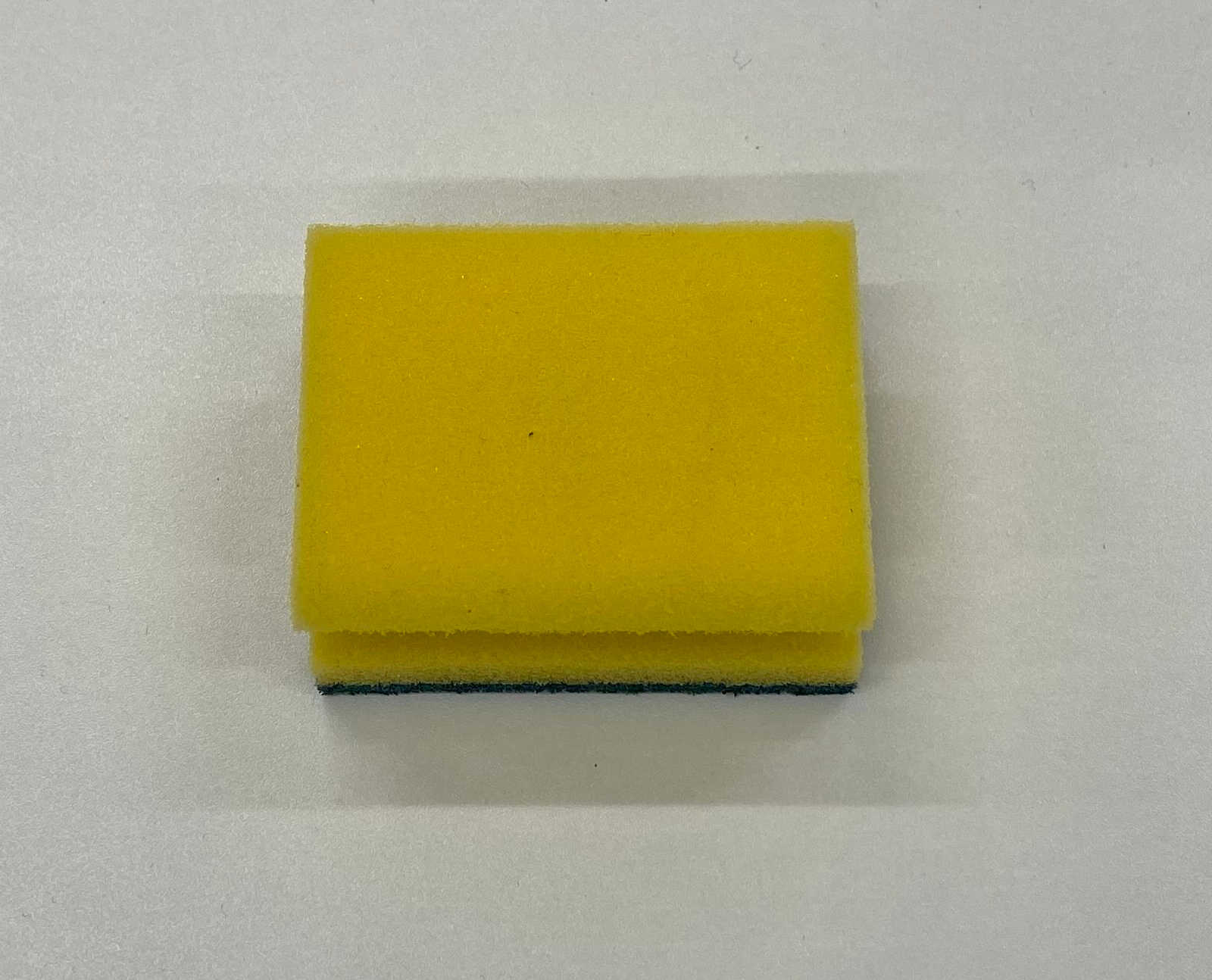}
		\caption{\label{fig:Obj_3} Object 3}
	\end{subfigure}
	~~~
	\begin{subfigure}[b]{0.268\columnwidth}
		\centering
		\includegraphics[width=\textwidth]{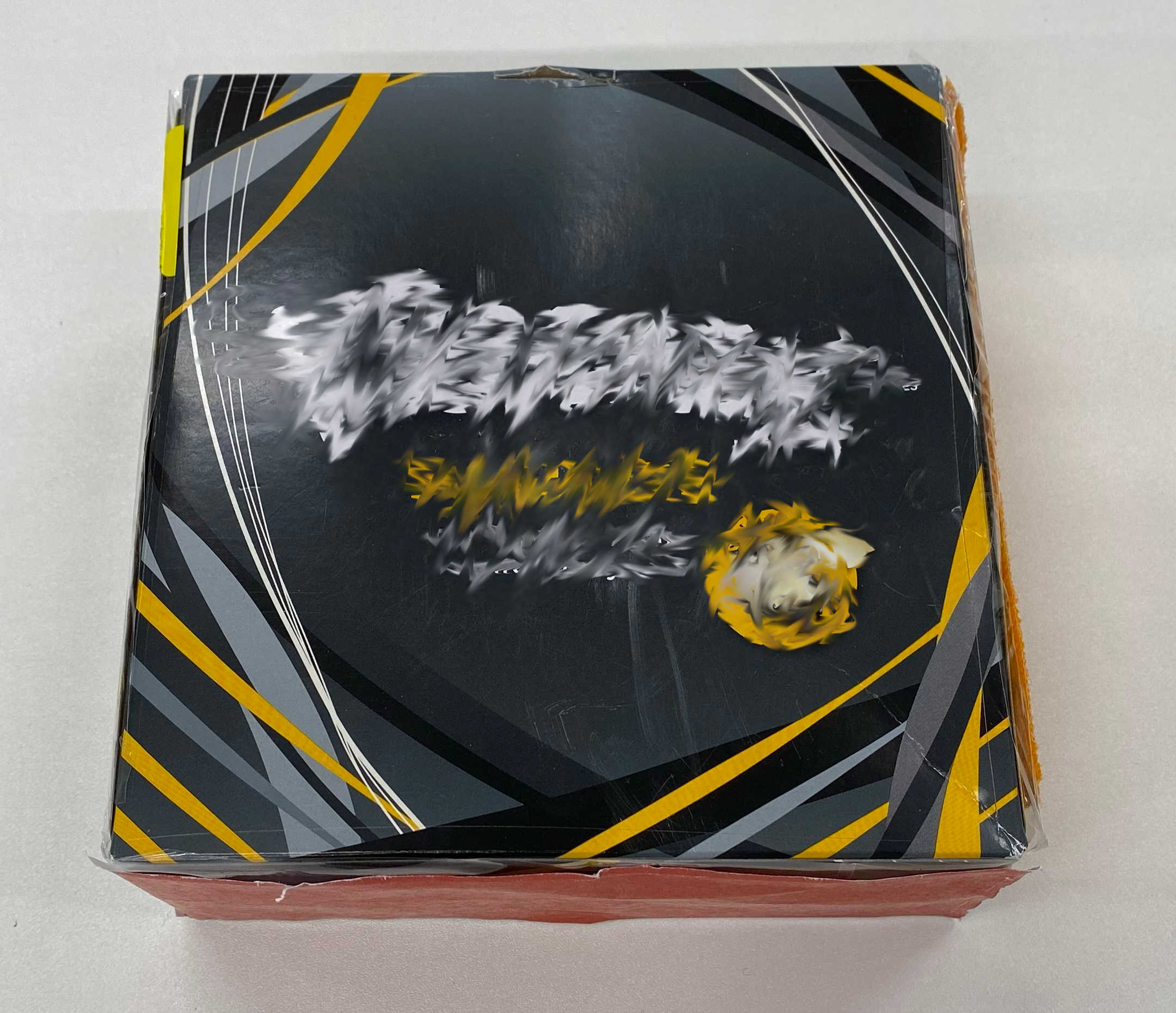}
		\caption{\label{fig:Obj_4} Object 4}
	\end{subfigure}
	~~~
	\begin{subfigure}[b]{0.3\columnwidth}
		\centering
		\includegraphics[width=\textwidth]{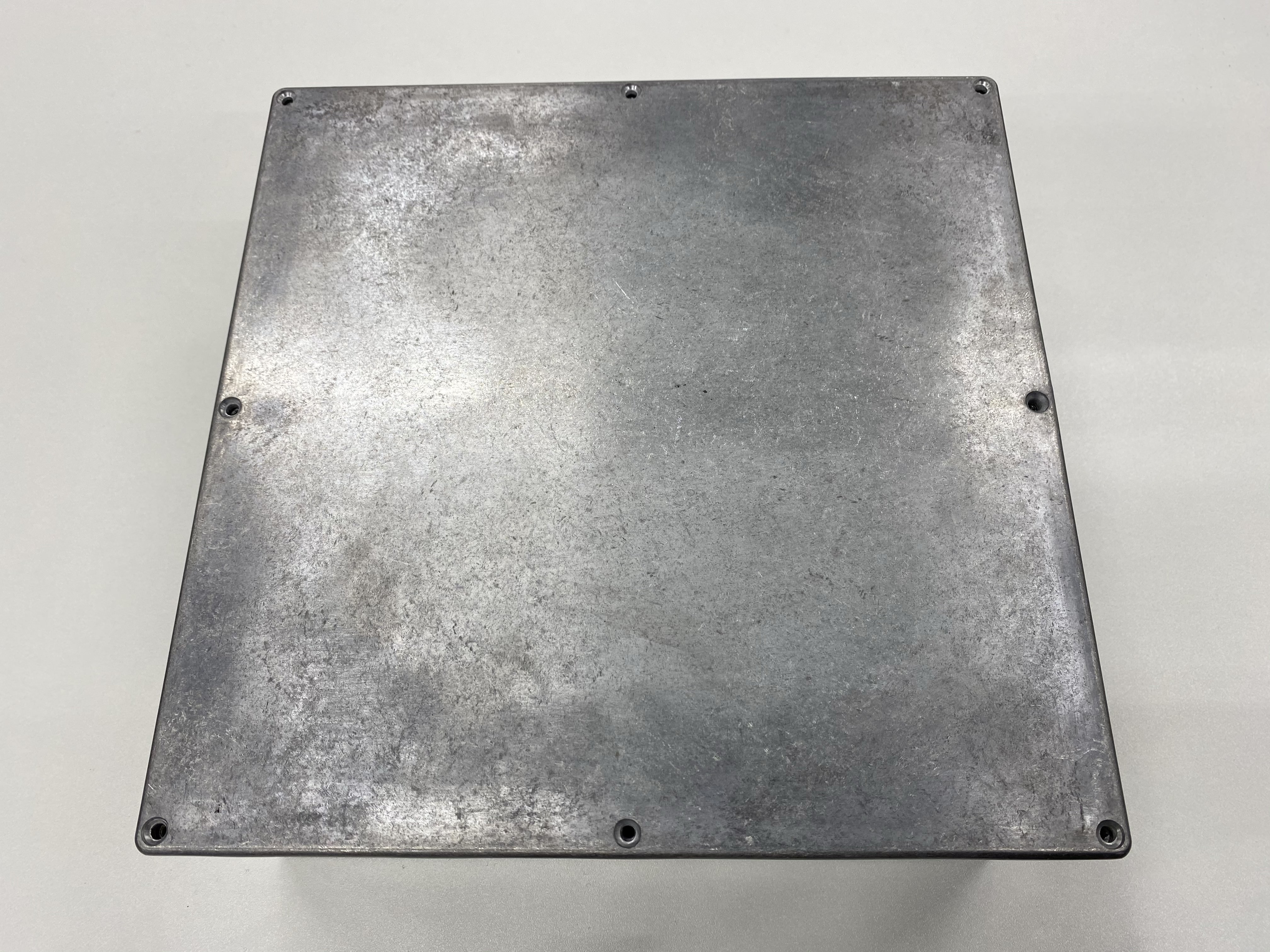}
		\caption{\label{fig:Obj_5} Object 5}
		
	\end{subfigure}
\caption{(a) : The lid of the plastic box shown in \autoref{fig:bilateralTelemanipulationSetup}. (b) Small rock. (c) Kitchen sponge (d) Tissue paper box (e)  Metal box.}
\label{fig:objects_for_interaction}
\end{figure}
\begin{enumerate}[i)]
    \item \textit{System impulse response}: 14 force impulses are applied to the replica's end-effector using a hammer while the master device is not held. The Matlab2020a (Matworks Inc, US) system identification toolbox is used to derive the transfer functions using the interaction force applied at the replica's end-effector as the system input and the velocity produced at the master as its output (i.e., $T(S)=V(s)/F(s)$).
    \item \textit{Interaction effects on system response}: An expert operator uses the master device to drive the replica robot to contact each of the 5 objects depicted in \autoref{fig:objects_for_interaction} before releasing hold of the master device. The selected objects' dynamics varies from stiff to compliant to evaluate the controllers' robustness to environmental interaction coupling effects.
    \item \textit{Dexterity comparison in challenging tasks}: we test the replica controlled by both IC and FIC, in a sequence of dexterous tasks to evaluate their performances. The tasks are conducted by a skilled operator, which is standard practice in the validation of teleoperation frameworks in robotics \cite{bilateralTeleManWithTimeDelays,ferraguti2013tank,ferraguti2015energy,minelli2019energy}. The operator is allowed a single attempt for each task and the choice of the starting controller was alternated to mitigate the operator bias. The tasks are: 1) Removing the lid from the metal box in \autoref{fig:Obj_5}, 2) Pushing a golf ball to an end-goal on a rough terrain (with sand, gravel, and rocks), 3) Driving a pile through the sand, 4) Diving the sponge in \autoref{fig:Obj_3} through a course, and 5) Press an E-stop button on a control panel.
    \item \textit{Robustness to Low-Bandwidth Feedback and Time-Delay}: We simulate a time-delay between master and replica via an adjustable signal buffer, while an adjustable zero-order hold is used to vary communication bandwidth. The affected signals are feedback force $F_\text{FB}$ provided as input to the master, the virtual force used to drive the replica's end-effector ($f_\text{v}$), and the desired pose $x_\text{d}$ provided as input to the $\text{FIC}_\text{R}$.
     We run an experiment where the operator needs to drive the replica to press two E-stop buttons placed \SI{10}{\centi\metre} apart. We test with both an expert and a naive operator to verify if system stability is independent from operator skill. The following communication conditions between master and replica are tested: 1) Signal sampling frequencies of $\SI{1}{k\hertz}$, $\SI{100}{\hertz}$ and $\SI{10}{\hertz}$,  2) Signal time delays of $\SI{0}{\second}$, $\SI{500}{m\second}$ and $\SI{1}{\second}$, and 3) combinations of both low-bandwidth feedback and time-delay. 
\end{enumerate}

\section{Results} \label{sec:Results}
\subsection {System impulsive response}
\begin{figure}[b]
    \centering
    \includegraphics[width=\columnwidth, trim= 5.5cm 10.5cm 5.5cm 10.5cm, clip]{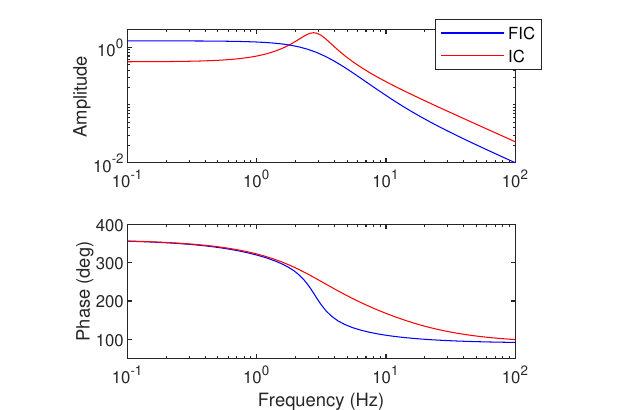}
    \caption{The impulse frequency response shows that the proposed controller and a traditional impedance controller have similar cut off frequencies at about $\SI{2}{\hertz}$, which is expected since the FIC share constant component of its impedance with the IC. However, the proposed method behaves as a critically damped system, while the traditional impedance controller is under-damped. }
    \label{fig:HzImpulses}
\end{figure}

The impulsive response of the two controllers  (\autoref{fig:HzImpulses}) show a system cut-off frequency of about 2Hz in both controllers.  
Although IC has 8x higher damping, it demonstrates an under-damped behavior. On the other hand, FIC's  higher overall stiffness (with lower damping), coupled with the fractal attractor's autonomous trajectories, allows it to maintain a desirable critically damped behavior. 

\subsection {Interaction effect on system response}
\begin{figure}[h]
    \centering
    \includegraphics[width=\columnwidth, trim=5.5cm 10cm 5.5cm 10cm, clip]{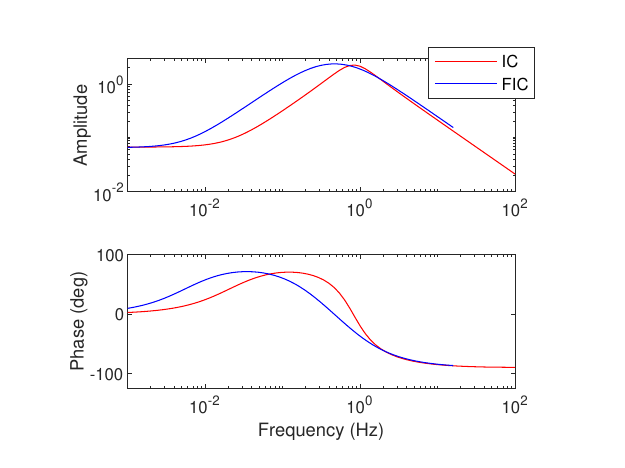}
    \caption{The spectral analysis of the teleoperation setup for the interaction with the 5 objects in \autoref{fig:objects_for_interaction} shows that the proposed method has a broader band-pass behaviour, centred at a frequency at about $0.5 \SI{}{\hertz}$ frequency. The main difference between the two controllers is that the IC peak is related to the underdamped behaviour of the system and it is higher than the band-pass characteristics of the system. On the other hand, the FIC is related to system band-pass behaviour, as can be observed in \autoref{fig:HzImpulses} for both controllers. }
    \label{fig:Hz5ObjTot}
\end{figure}

\begin{figure}[h]
    \centering
    \begin{subfigure}[b]{\columnwidth}
        \centering 
        \includegraphics[width=\columnwidth, trim= 5.5cm 10.5cm 5.5cm 10.5cm, clip]{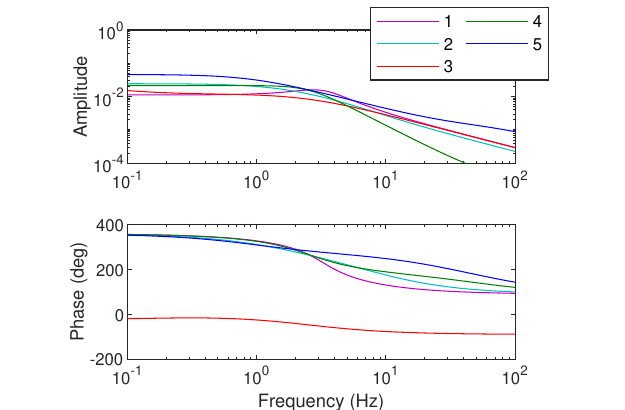}
        \caption{ \label{fig:FIC5Objs}}
    \end{subfigure}
        \begin{subfigure}[b]{\columnwidth}
        \centering
        \includegraphics[width=\columnwidth, trim= 5.5cm 10.5cm 5.5cm 10.5cm, clip]{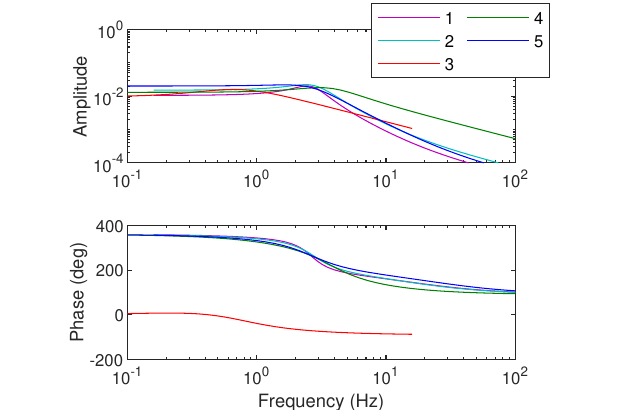}
        \caption{ \label{fig:IC5Objs}}
    \end{subfigure}
\caption{Separating the data in \autoref{fig:Hz5ObjTot} based on the object interaction, we have performed system identification for each task. (a) describes the FIC spectra. (b) describes the IC spectra. The FIC plots show that its higher band-pass is due to its synchronisation with the environment as shown by the significant variance in the phase diagram. In contrast the IC shows a similar behaviour with all the objects.}
\label{fig:Hz5ObjsSep}
 \end{figure}

\autoref{fig:Hz5ObjTot} also confirms these characteristics when interacting with objects. However, we also observe a band-pass behaviour due to the user driving the robot via the master device (at least until the replica's end-effector makes contact with the object), which filters the lower frequencies. 
It can also be observed how the band-pass is wider where the replica is driven via FIC controller.
\autoref{fig:Hz5ObjsSep} compares the behaviour of the two controllers when interacting with different objects. The data show that the transient behaviour of the proposed method is more variable, and it has a lower attenuation of the input signal. 
 
\subsection {Dexterity comparison between the IC and the FIC}
The completion times for the 5 tasks performed by an expert operator (using high quality communication between master and replica) are, respectively, $\SI{34}{\second}$, $\SI{26}{\second}$, $\SI{22}{\second}$, $\SI{94}{\second}$ and $\SI{6}{\second}$ for the FIC. The times when using IC are NA (task failed), $\SI{32}{\second}$, NA (task failed), 
$\SI{152}{\second}$ and $\SI{32}{\second}$, respectively. The data show that the operator is faster in all tasks when executing with the FIC controller.  Task \text{1} failed with IC due to the controller command triggering the robot safety mechanisms. Task \text{3} also failed with IC due to lack of dexterity. These experiments confirm that the FIC enables better interaction dexterity compared to the IC controller, as highlighted in the attached video. 

\begin{figure}[H]
\centering
    \begin{subfigure}[b]{0.9\columnwidth}
	    \centering
		\includegraphics[width=\columnwidth]{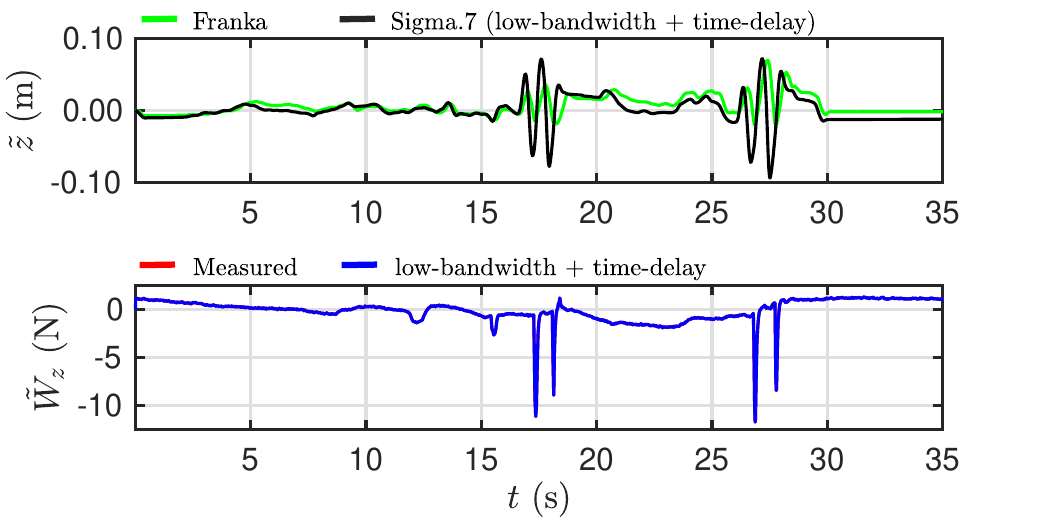}
		\caption{Expert User: $f = \SI{1}{k\hertz}, t_{\text{delay}} = \SI{0}{\second}$\label{fig:exp_1_f_1000Hz_t_0s}}
	\end{subfigure}
	\begin{subfigure}[b]{0.9\columnwidth}
		\centering
		\includegraphics[width=\columnwidth]{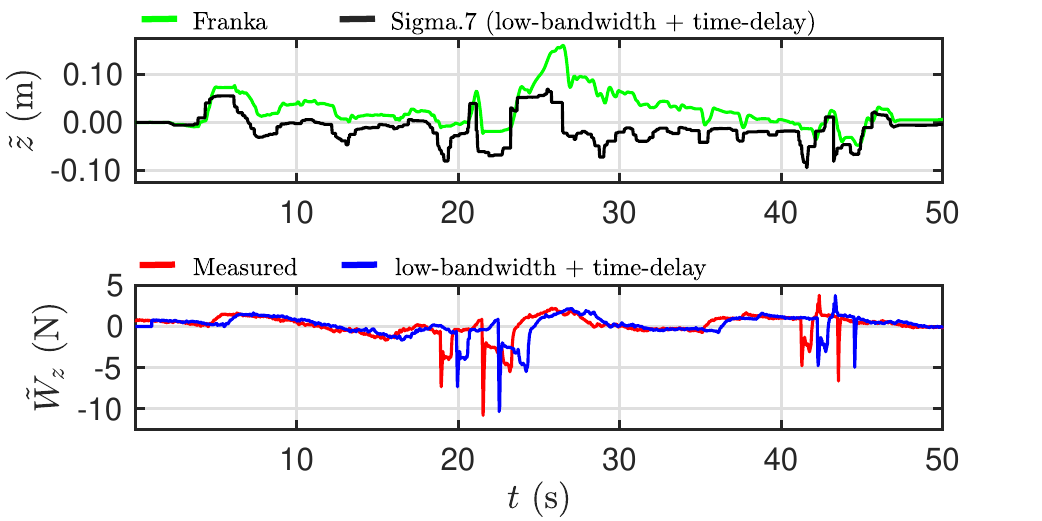}
		\caption{Expert User: $f = \SI{10}{\hertz}, t_{\text{delay}} = \SI{1}{\second}$\label{fig:exp_3_f_10Hz_t_1s}}
	\end{subfigure}
\caption{E-Stop button pressing experiments of the expert operator, under ideal (a) and poor (b) quality signal transmission. Positions of both master and replica ($\tilde{z}$) are shown with environmental force feedback (${\tilde{W}}_z$).  The expert can more easily adapt to challenging conditions to complete the task. \label{Expert_User}}
\label{fig:objects_for_interaction_expert}
\end{figure}

\begin{figure}[h]
\centering
    \begin{subfigure}[b]{0.9\columnwidth}
	    \centering
		\includegraphics[width=\columnwidth]{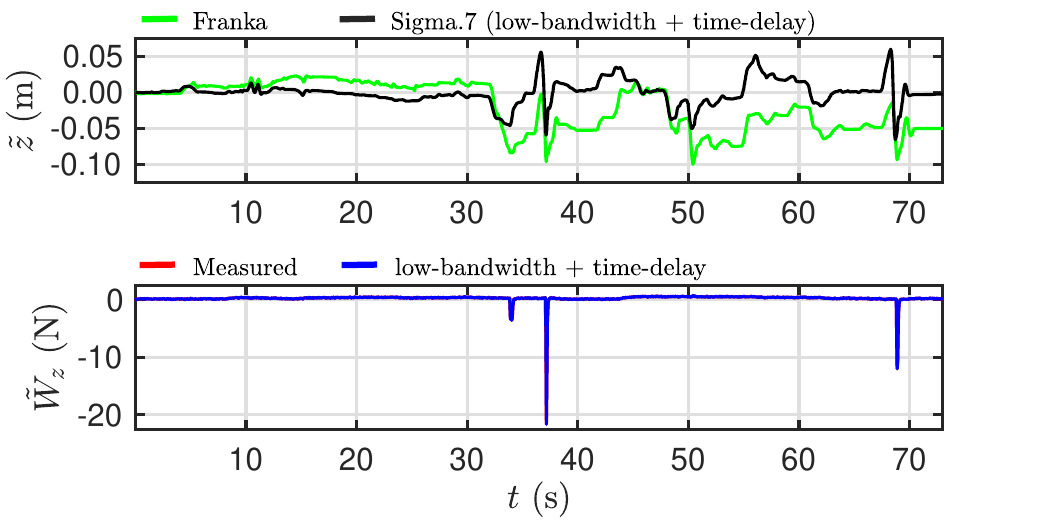}
		\caption{Naive User: $f = \SI{1}{k\hertz}, t_{\text{delay}} = \SI{0}{\second}$\label{fig:exp_1_dan_f_1000Hz_t_0s}}
	\end{subfigure}
	\begin{subfigure}[b]{0.9\columnwidth}
		\centering
		\includegraphics[width=\columnwidth]{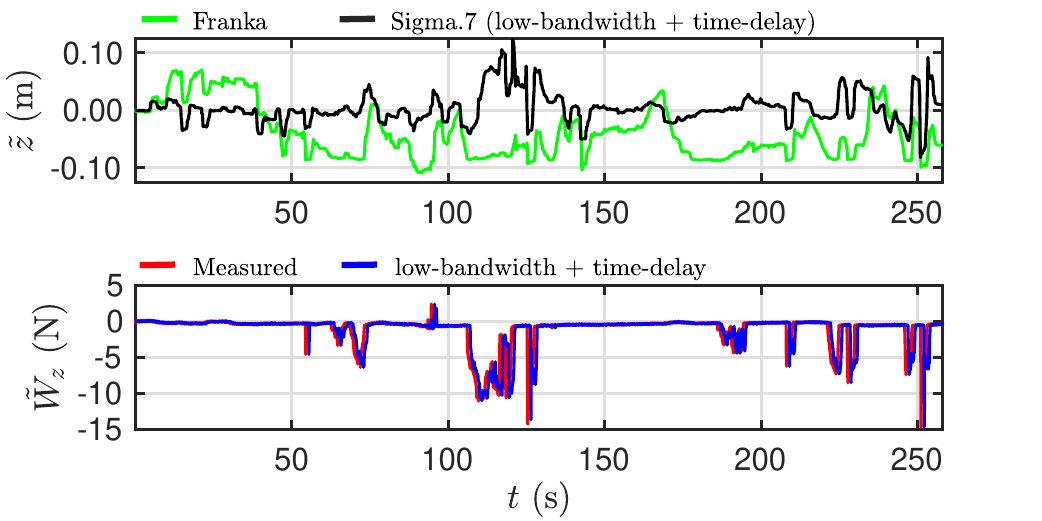}
		\caption{Naive User: $f = \SI{10}{\hertz}, t_{\text{delay}} = \SI{1}{\second}$\label{fig:exp_3_dan_f_10Hz_t_1s}}
	\end{subfigure}
\caption{E-Stop button pressing experiments of the naive operator, under ideal (a) and poor (b) signal transmission. Although it is more difficult for the naive user to adapt to degraded transmission (resulting in a substantial increase in execution time), the system never became unstable and the operator was able to achieve the task. \label{Naive_User}}
\label{fig:objects_for_interaction_naive}
\end{figure}

\subsection {Robustness to Low-Bandwidth Feedback and Time-Delay}
\autoref{Expert_User} and \autoref{Naive_User} show trajectories recorded for a button pushing task from the expert and naive operators using our proposed method under high quality ($f=\SI{1}{\kilo \hertz}$  and $t_\text{delay}=\SI{0}{\second}$) and poor ($f=\SI{10}{\hertz}$  and $t_\text{delay}=\SI{1}{\second}$) signal transmission. Comparing the data we can observe that, as expected, the expert user has more confident movement and better control over the vertical interaction force $W_\text{z}$. Nonetheless, the naive subject can successfully complete the assigned task without triggering the robot safety mechanisms in any of the analysed cases. The completion times are reported in 
\autoref{NaiveExpertComparison} for all tested conditions. They indicate that the expert user is between 2 and 4 times faster than the naive user, but both are able to complete all tasks successfully. Thus, we can conclude that the proposed method is robust to extreme delay and low communication bandwidth, and moreover, controller stability is independent of the operator skill.

\section{DISCUSSION} \label{sec:Discussion} 
We have presented a passive telemanipulation framework based on the Fractal Impedance Control (FIC) which guarantees stability in the presence of delayed and reduced feedback bandwidth. Our results validate that the proposed  controller is robust to dynamic interaction with external environments as well as to extremely poor signal transmission between master and replica.  Spectral analysis shown in \autoref{fig:Hz5ObjTot} demonstrates that the proposed method has a wider band-pass and lower attenuation of inputs when driven by a user. Thus our system enables higher transparency for telemanipulation tasks, compared to the traditional impedance controller. 

In terms of delays and reduced bandwidth in the communication, the analysis of the controller performance indicates that our method can guarantee stability of interaction during manipulation even in extreme conditions ($f=\SI{10}{\hertz}$ and $t_\text{delay}=\SI{1}{\second}$) without requiring an expert operator. The outcome of the dexterity comparison between FIC and the IC controllers shows that the proposed method allows better maneuvering and has more robust interaction than a traditional impedance controller. 

In conclusion, the results show that the proposed method has intrinsic adaptability to different environmental dynamics, even without performing online gain tuning. Robustness to low-bandwidth and time-delayed signals shows its promise for applications in extreme environments such as planetary space exploration. Future work will focus on extending the controller on bi-manual tasks, drilling, cutting and surface finishing.

\begin{figure}[h]
\centering
    \begin{subfigure}[b]{\columnwidth}
		\centering
		 \vspace*{+2mm}
		\includegraphics[width=\textwidth]{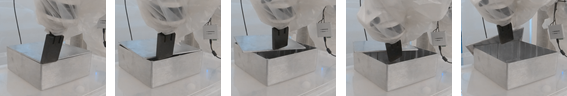}
		\caption{Task 1 \label{fig:Ex1}}
	\end{subfigure}
    \begin{subfigure}[b]{\columnwidth}
		\centering
		\includegraphics[width=\textwidth]{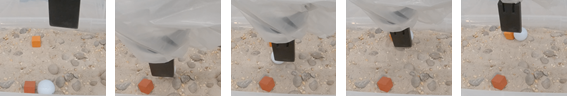}
		\caption{Task 2 \label{fig:Ex2}}
	\end{subfigure}
	\begin{subfigure}[b]{\columnwidth}
		\centering
		\includegraphics[width=\textwidth]{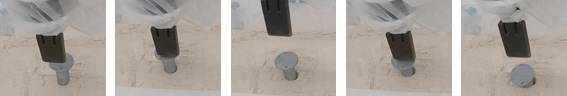}
		\caption{Task 3 \label{fig:Ex3}}
	\end{subfigure}
	\begin{subfigure}[b]{\columnwidth}
		\centering
		\includegraphics[width=\textwidth]{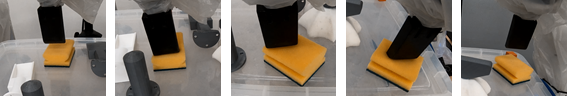}
		\caption{Task 4 \label{fig:Ex4}}
	\end{subfigure}
	\begin{subfigure}[b]{\columnwidth}
		\centering
		\includegraphics[width=\textwidth]{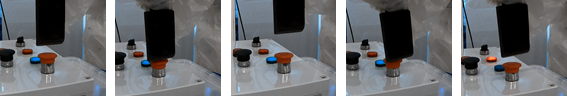}
		\caption{Task 5 \label{fig:Ex5}}
	\end{subfigure}
\caption{Snapshots of the 5 tasks performed by expert user using the FIC. }
\label{fig:carried_out_exps}
\end{figure}

\begin{table}[!htb]
\caption{Expert and Naive User Task Completion Times $\SI{}{(\second)}$}	
    \begin{center}
	    \begin{tabular}{| c | c | c|}
		\hline 
		Communication Condition & Expert User & Naive User \\ \hline
		$f = \SI{1}{k\hertz}$, $t_{\text{delay}} = \SI{0}{\second}$ & 31 & 69 \\ \hline
		$f = \SI{100}{\hertz}$, $t_{\text{delay}} = \SI{0}{\second}$ & 39 & 73 \\ \hline
		$f = \SI{10}{\hertz}$, $t_{\text{delay}} = \SI{0}{\second}$ & 41 & 99 \\ \hline
		$f = \SI{1}{k\hertz}$, $t_{\text{delay}} = \SI{500}{m\second}$ & 49 & 105 \\ \hline
		$f = \SI{100}{\hertz}$, $t_{\text{delay}} = \SI{500}{m\second}$ & 51 & 209 \\ \hline
		$f = \SI{10}{\hertz}$, $t_{\text{delay}} = \SI{500}{m\second}$ & 57 & 241 \\ \hline
		$f = \SI{1}{k\hertz}$, $t_{\text{delay}} = \SI{1}{\second}$ & 61 & 243 \\ \hline
		$f = \SI{100}{\hertz}$, $t_{\text{delay}} = \SI{1}{\second}$ & 63 & 247 \\ \hline
		$f = \SI{10}{\hertz}$, $t_{\text{delay}} = \SI{1}{\second}$ & 66 & 258 \\ \hline
		\end{tabular}
	\end{center}
	\label{NaiveExpertComparison}
\label{table3}
\end{table}

\bibliography{FIC_in_teleoperation_bibtex_file}
\bibliographystyle{IEEEtran}
\balance

\end{document}